\documentclass[journal]{IEEEtran}
\usepackage{bm}

\newcommand{\TrainingSet}{\bm{T}} 
\newcommand{\ValidationSet}{\bm{V}} 
\newcommand{\TestSet}{\Psi}

\usepackage{amsfonts}
\usepackage{graphicx}
\usepackage{makecell}
\usepackage{tikz}
\usepackage{collcell}
\usepackage{colortbl}
\usepackage{pgfplots}
\usepackage{pgfplotstable}
\usepackage{multirow}
\usepackage{url}
\usepackage[all]{xy}
\usepackage{cite}

\ifCLASSINFOpdf
\else
\fi

\pgfplotstableset{
    /color cells/min/.initial=0,
    /color cells/max/.initial=1000,
    /color cells/textcolor/.initial=,
    color cells/.code={%
        \pgfqkeys{/color cells}{#1}%
        \pgfkeysalso{%
            postproc cell content/.code={%
                \begingroup
                \pgfkeysgetvalue{/pgfplots/table/@preprocessed cell content}\value
\ifx\value\empty
\endgroup
\else
                \pgfmathfloatparsenumber{\value}%
                \pgfmathfloattofixed{\pgfmathresult}%
                \let\value=\pgfmathresult
                \pgfplotscolormapaccess
                    [\pgfkeysvalueof{/color cells/min}:\pgfkeysvalueof{/color cells/max}]%
                    {\value}%
                    {\pgfkeysvalueof{/pgfplots/colormap name}}%
                \pgfkeysgetvalue{/pgfplots/table/@cell content}\typesetvalue
                \pgfkeysgetvalue{/color cells/textcolor}\textcolorvalue
                \toks0=\expandafter{\typesetvalue}%
                \xdef\temp{%
                    \noexpand\pgfkeysalso{%
                        @cell content={%
                            \noexpand\cellcolor[rgb]{\pgfmathresult}%
                            \noexpand\definecolor{mapped color}{rgb}{\pgfmathresult}%
                            \ifx\textcolorvalue\empty
                            \else
                                \noexpand\color{\textcolorvalue}%
                            \fi
                            \the\toks0 %
                        }%
                    }%
                }%
                \endgroup
                \temp
\fi
            }%
        }%
    }
}

\begin{document}
\title{Validating Hyperspectral Image Segmentation}
\author{Jakub Nalepa,~\IEEEmembership{Member,~IEEE}, 
        Michal Myller,
        and Michal Kawulok,~\IEEEmembership{Member,~IEEE}
\thanks{This work was funded by European Space Agency (HYPERNET project).}
\thanks{J.~Nalepa and M.~Kawulok are with Silesian University of Technology, Gliwice, Poland (e-mail: \{jakub.nalepa, michal.kawulok\}@polsl.pl).}
\thanks{J.~Nalepa, M.~Myller, and M.~Kawulok are with KP Labs, Gliwice, Poland (e-mail \{jnalepa, mmyller, mkawulok\}@kplabs.pl).}
}


\maketitle
\begin{abstract}
Hyperspectral satellite imaging attracts enormous research attention in the remote sensing community, hence automated approaches for precise segmentation of such imagery are being rapidly developed. In this letter, we share our observations on the strategy for validating hyperspectral image segmentation algorithms currently followed in the literature, and show that it can lead to over-optimistic experimental insights. We introduce a new routine for generating segmentation benchmarks, and use it to elaborate ready-to-use hyperspectral training-test data partitions. They can be utilized for fair validation of new and existing algorithms without any training-test data leakage.
\end{abstract}


\IEEEpeerreviewmaketitle

\section{Introduction} \label{sec:intro}

With the current sensor advancements, hyperspectral satellite imaging (HSI) is becoming a mature technology which captures a very detailed spectrum (usually more than a hundred of bands) of light for each pixel. Such a big amount of reflectance information about the underlying material can help in accurate HSI \emph{segmentation} (deciding on the boundaries of objects of a given class). Hence, HSI is being actively used in a range of areas, including precision agriculture, military, surveillance, and more~\cite{Dundar2018}. The state-of-the-art HSI
segmentation methods include conventional machine-learning
algorithms, which can be further divided into unsupervised~\cite{TARABALKA20102367,Bilgin2011} and supervised~\cite{Dundar2018,Amini2018,Li2018} techniques, and modern deep-learning (DL) algorithms~\cite{Li2014,Chen2015,Chen2014,Zhao2016,Chen2016,Liu2017,Zhong2017,Mou2017,Santara2017,Lee2017,Gao_2018,Ribalta2018} that do not require feature engineering. DL techniques allow for extracting \emph{spectral} features (e.g.,~using deep belief networks [DBNs]~\cite{Liu2017,Zhong2017} and recurrent neural networks [RNNs]~\cite{Mou2017}) or both \emph{spectral} and \emph{spatial} pixel's information---mainly using convolutional neural networks (CNNs)~\cite{Zhao2016,Chen2016,Santara2017,Lee2017,Gao_2018,Ribalta2018} and some DBNs~\cite{Li2014,Chen2015}.

\begin{table}[ht!]
	\scriptsize
	\centering
	\caption{A summary of the HSI benchmarks and experimental settings used to validate the state-of-the-art segmentation techniques.}
	\label{tab:literature_review}
	\renewcommand{\tabcolsep}{0.21cm}
	\begin{tabular}{lrll}
		\Xhline{2\arrayrulewidth}
		Method & Year & Datasets$^*$ & Settings\\
		\hline
Multiscale superpixels \cite{Dundar2018} & 2018 & IP, PU & Random\\ 
Watershed + SVM \cite{TARABALKA20102367} & 2010 & PU & Arbitrary\\ 
Clustering (SVM) \cite{Bilgin2011} & 2011 & Washington DC, PU & Full image\\ 
Multiresolution segm. \cite{Amini2018} & 2018 & Three in-house datasets & Random\\ 
Region expansion \cite{Li2018} & 2018&  PU, Sa, KSC & Full image\\ 
DBN (spatial-spectral) \cite{Li2014} & 2014 & HU & Random\\ 
DBN (spatial-spectral) \cite{Chen2015} & 2015 & IP, PU & Random\\ 
Deep autoencoder \cite{Chen2014} & 2014 & KSC, PU & Monte Carlo\\ 
CNN \cite{Zhao2016} & 2016 & PC, PU, & Random\\ 
CNN \cite{Chen2016} & 2016& IP, PU, KSC & Monte Carlo\\ 
Active learning + DBN \cite{Liu2017} & 2017 & PC, PU, Bo & Random\\ 
DBN (spectral) \cite{Zhong2017} & 2017 &  IP, PU & Random\\ 
RNN (spectral) \cite{Mou2017} & 2017 & PU, HU, IP & Random\\ 
CNN \cite{Santara2017} & 2017 & IP, Sa, PU & Random\\ 
CNN \cite{Lee2017} & 2017 & IP, Sa, PU & Monte Carlo\\ 
CNN \cite{Gao_2018} & 2018 & IP, Sa, PU & Monte Carlo\\ 
CNN \cite{Ribalta2018} & 2018 & Sa, PU & Random\\

		\Xhline{2\arrayrulewidth}
\multicolumn{4}{p{8cm}}{$^*$ IP---Indian Pines; PU---Pavia University; Sa---Salinas; KSC---Kennedy Space Center; HU---Houston University; PC---Pavia Centre; Bo---Botswana}
	\end{tabular}
\end{table}

Validation of the HSI segmentation algorithms is a challenging task, especially due to the limited number of manually annotated ground-truth sets. Virtually all segmentation techniques have been tested using up to three HSI benchmarks (Table~\ref{tab:literature_review}), with the Pavia University, Indian Pines, and Salinas Valley images being the most popular (exploited in 15, 8, and 5 out of 17 recently published papers inspected in this letter). A common approach is to extract \emph{training} $\TrainingSet$, and \emph{test} $\TestSet$ (used for quantifying the generalization of the trained model), and possibly \emph{validation} $\ValidationSet$ (used for hyper-parameter selection or guiding the training process in deep-learning approaches) subsets from \emph{the very same} HSI. Almost all analyzed algorithms were validated using random splits (possibly drawn multiple times in the Monte-Carlo setting), in which pixels from an input HSI are selected as training and test pixels at random (without overlaps). Unfortunately, different authors report different divisions (in terms of the percentages of pixels taken as training and test ones) which makes fair comparison of new and existing techniques troublesome.

\begin{figure}[ht!]
	\centering
	\includegraphics[width=0.75\columnwidth]{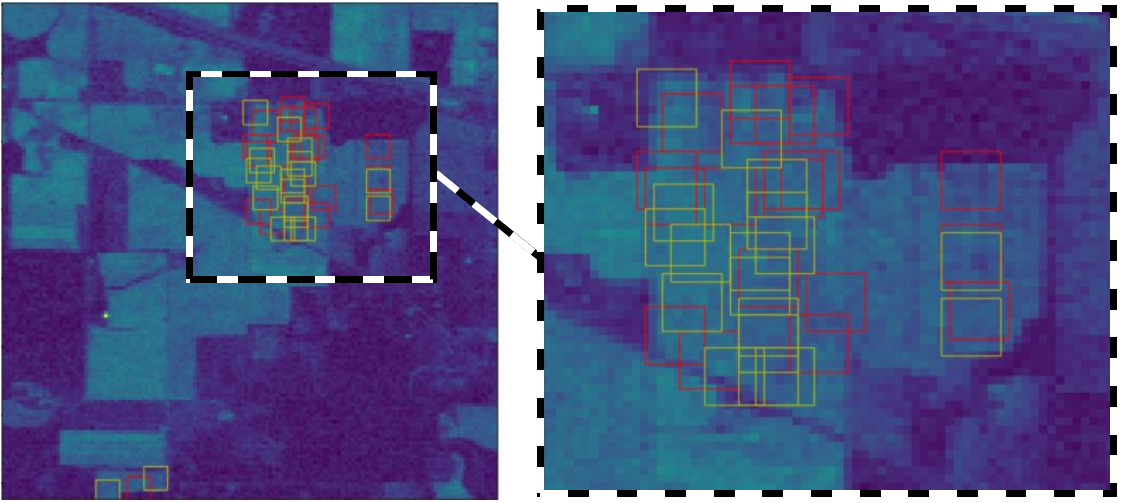}
	\caption{Random pixel selection from Indian Pines to the training and test sets (overlapping yellow and red squares) can cause a leak of information.}
	\label{fig:datasets1}
\end{figure}

\begin{figure}[ht!]
	\centering
\renewcommand{\tabcolsep}{0.0cm}


\includegraphics[width=0.75\columnwidth]{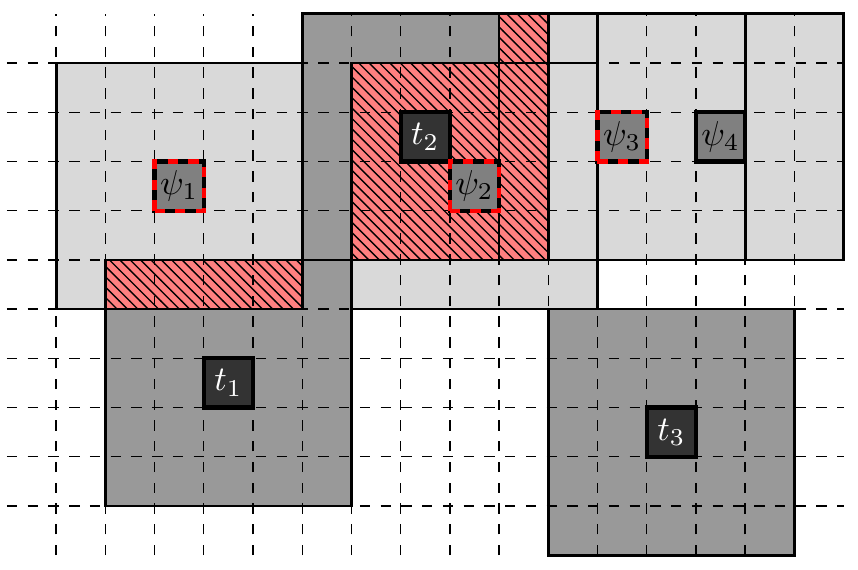}

	\caption{Training ($t_i$) and test ($\psi_i$) pixels with their spatial neighborhoods. The overlapping pixels (red-shadowed area), are ``leaked'' across these sets.}
	\label{fig:example_data_leakage}
\end{figure}

Random selection of $\TrainingSet$ and $\TestSet$ can very easily lead to over-estimating the performance of the validated HSI segmentation methods, if the features describing a single pixel are extracted from its neighborhood in the spatial domain. This problem is illustrated for Indian Pines in Fig.~\ref{fig:datasets1} and explained in Fig.~\ref{fig:example_data_leakage}---three pixels ($t_i$) are selected to $\TrainingSet$
and four pixels ($\psi_i$) are in $\TestSet$. It can be seen that for such random selection, the $5\times5$ neighborhoods of the pixels in $\TrainingSet$ and $\TestSet$ overlap. As a result, some of the pixels that are used for testing, have been seen during training (marked as red), and this even may concern the very pixels in $\TestSet$ (here, $\psi_2$). In this example, only $\psi_4$ is independent from $\TrainingSet$. The most commonly reported scenario is to select the pixels to $\TrainingSet$, while the remaining pixels form $\TestSet$. In such cases, all the pixels which are not in $\TrainingSet$, but are in the neighborhood of the pixels from $\TrainingSet$, are used for evaluation, resulting in information leak between training and test subsets.

This observation may appear trivial, however in practically all works which extract spatial-spectral features, it is not reported to exclude entire neighborhoods of the pixels used for training from $\TestSet$. This problem is even more apparent for deep CNNs, where the features used for classifying a given pixel $p$ are extracted from an entire patch, whose central pixel is $p$. Although in the papers mentioned here, the dimensions of these patches do not exceed $10\times 10$ pixels, it is not uncommon to see patches larger than $100\times 100$, and some architectures, including a famous U-Net~\cite{RonnebergerFB15}, can process an entire image. Such networks are capable of learning high-level rules, which may be image-specific, leading to poor generalization, if derived from a single image. For example, a CNN may learn that a broccoli field is located alongside a bitumen road, in addition to some spectral and local features. Moreover, CNNs require lots of data in $\TrainingSet$, so their performance is validated using cross-validation with $\TrainingSet$ much larger than $\TestSet$, hence it may be highly possible that all the pixels in $\TestSet$ are seen during training. Overall, evaluation of HSI segmentation, especially performed using CNNs, may render over-optimistic conclusions.

\begin{figure*}[ht!]
\centering
\includegraphics[width=0.7\paperwidth]{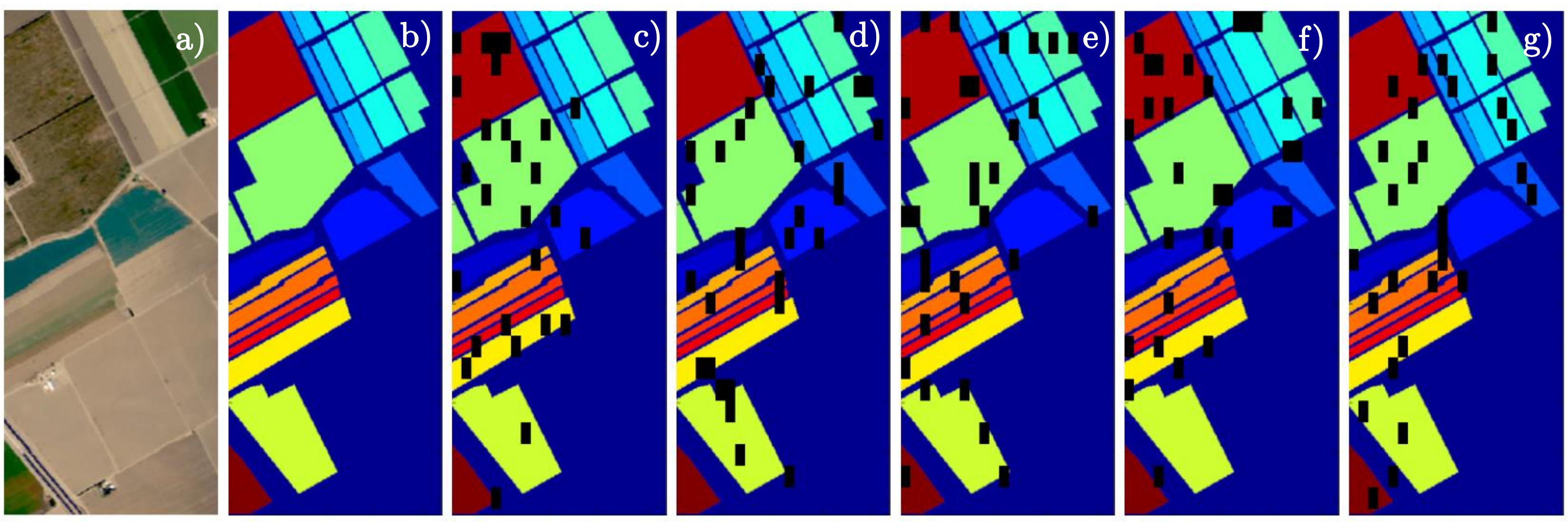}
	\caption{Our benchmark data (five non-overlapping folds) generated over the Salinas Valley set: a) true-color composite, b) ground-truth, c)--g) visualization of all folds: black patches contain training pixels, whereas the other pixels are taken as test data.}
	\label{fig:salinas_folds}
\end{figure*}

\begin{figure*}[ht!]
\centering
\includegraphics[width=0.7\paperwidth]{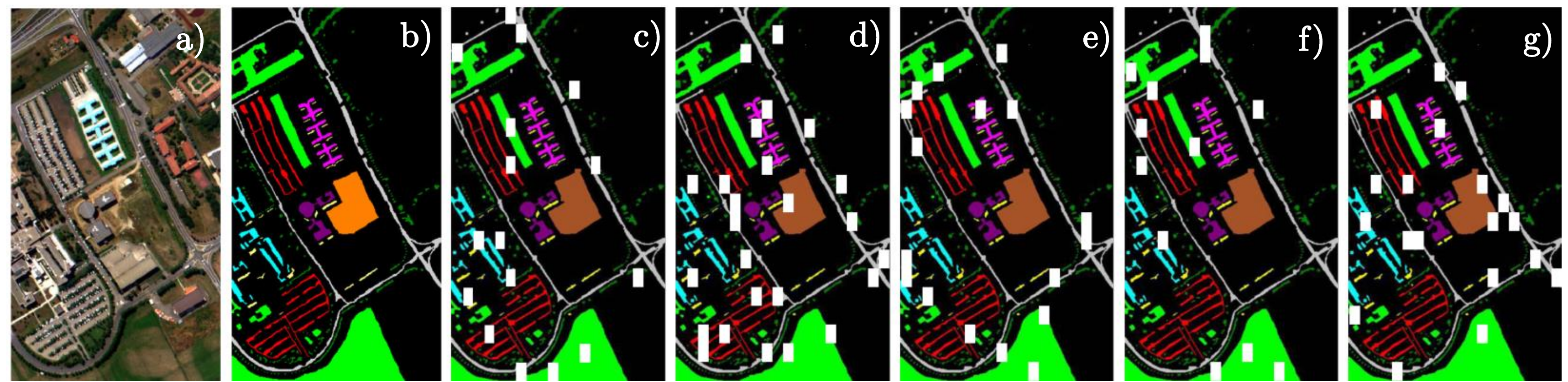}
	\caption{Our benchmark data (five non-overlapping folds) generated over the Pavia University set: a) true-color composite, b) ground-truth, c)--g) visualization of all folds: white patches contain training pixels, whereas the other pixels are taken as test data.}
	\label{fig:pavia_folds}
\end{figure*}

In this letter, we analyze a recent spatial-spectral method from the literature that exploits a CNN~\cite{Gao_2018}, alongside our spectral CNN which does \emph{not} utilize spatial information, and we report our experiments (Section~\ref{sec:experiments}) to demonstrate the potential consequences of such training-test information leakages. Our contribution lies in introducing a simple and flexible method for extracting training and test sets from an input HSI (Section~\ref{sec:method}) which does not suffer from the training-test data leakage. We use it to generate a benchmark (based on the Salinas Valley, Pavia University, and Indian Pines images) which can be utilized for fair comparison of HSI segmentation algorithms. The results obtained using these data clearly show that training-test information leakages, together with other problems including intrinsic correlation of the training data and imbalanced representation (which can even mean absence of some classes in $\TrainingSet$) can drastically decrease the estimated accuracy for widely-used benchmark scenes.

\section{Patch-Based HSI Segmentation Benchmarks}\label{sec:method}

In the proposed patch-based algorithm to generate HSI segmentation benchmarks, we randomly select a set of patches of size $w_p\times h_p$ from an input image, where $w_p$ and $h_p$ denote their width and height, respectively. The size is relative to the original image dimensions, therefore it is $t_w w_I \times t_h h_I$, where $w_I$ and $h_I$ are the width and height of the image, and $t_w$ and $t_h$ are given in \%. These patches are being drawn until at least $\left|\left|\TrainingSet\right|\right|$ training pixels are selected (this set may be balanced or imbalanced). Importantly, the size of a patch should be greater or equal to the size of a pixel neighborhood used for classification in a segmentation technique being validated. We treat the extracted patches as separate images (we remove these data from the original image), so the pixels from the patches cannot be seen during testing.


Using our technique, we created patch-based multi-class HSI benchmarks over three most-popular hyperspectral images: Salinas Valley (NASA Airborne Visible/Infrared Imaging Spectrometer AVIRIS sensor), Pavia University (Reflective Optics System Imaging Spectrometer ROSIS sensor), and Indian Pines (AVIRIS). AVIRIS registers 224 channels with wavelengths in a 400 to 2450 nm range (visible to near-infrared), with 10 nm bandwidth, and it is calibrated to within 1 nm. ROSIS collects the spectral radiance data in 115 channels in a 430 to 850 nm range (4 nm nominal bandwidth). All benchmarks (together with a summary on the number of training and test pixels in each fold) generated using our method, alongside its \texttt{Python} implementation are available at \url{https://tinyurl.com/ieee-grsl}. To verify the impact of such partitions on the performance of a state-of-the-art method exploiting spatial information, we created splits that contain similar numbers of training/test pixels as reported in~\cite{Gao_2018}. Additionally, our sets are imbalanced and may even not contain pixels from a given class (which reflects real-life satellite imaging scenarios where we cannot assume that all classes of interest appear within a scene). We discuss the considered HSI images in more detail below.

\subsubsection{Salinas Valley}

This set ($512\times 217$ pixels) was captured over Salinas Valley in California, USA, with a spatial resolution of 3.7 m. The image shows different sorts of vegetation (16 classes), and it contains 224 bands (20 are dominated by water absorption). We extract 5 folds (Fig.~\ref{fig:salinas_folds}), and $w_p=22$ and $ h_p=10$ (in pixels), hence $t_w=4.6\%$ and $t_h=4.2\%$.

\subsubsection{Pavia University}

This set ($610\times 340$) was captured over the Pavia University in Lombardy, Italy, with a spatial resolution of 1.3 m. The image shows an urban scenery (9 classes), and contains 103 channels (12 water absorption-dominated bands were removed). We extract 5 folds (Fig.~\ref{fig:pavia_folds}), $w_p=30$ and $ h_p=17$, hence $t_w=5\%$ and $t_h=4.9\%$.

\begin{figure}[ht!]
	\centering
\includegraphics[width=0.35\paperwidth]{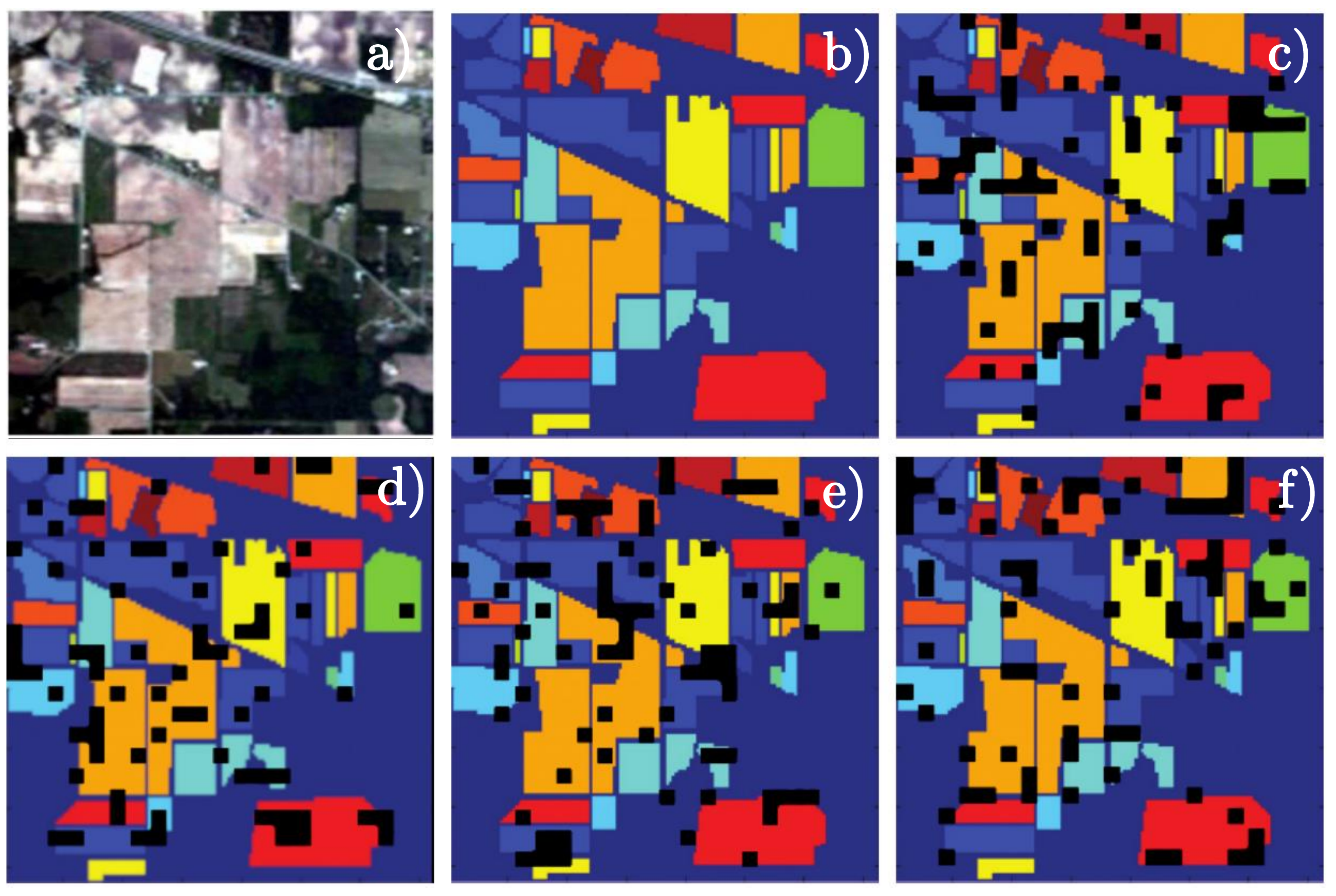}
	\caption{Our benchmark data (four non-overlapping folds) generated over the Indian Pines set: a) true-color composite, b) ground-truth, c)--f) visualization of all folds (meanings of colors as in Fig.~\ref{fig:salinas_folds}).}
	\label{fig:indiana_folds}
\end{figure}

\subsubsection{Indian Pines}

This set ($145\times 145$) was captured over the North-Western Indiana, USA, with a spatial resolution of 20~m. It shows agriculture and forest (or other natural perennial vegetation) areas (16 classes). The set contains 200 channels (24 water-dominated bands were removed). We extract 4 folds (Fig.~\ref{fig:indiana_folds}), $w_p=7$ and $ h_p=7$, hence $t_w=t_h=4.8\%$.

\section{Experiments}\label{sec:experiments}

\begin{figure}[ht!]
	\centering
	\includegraphics[width=0.82\columnwidth]{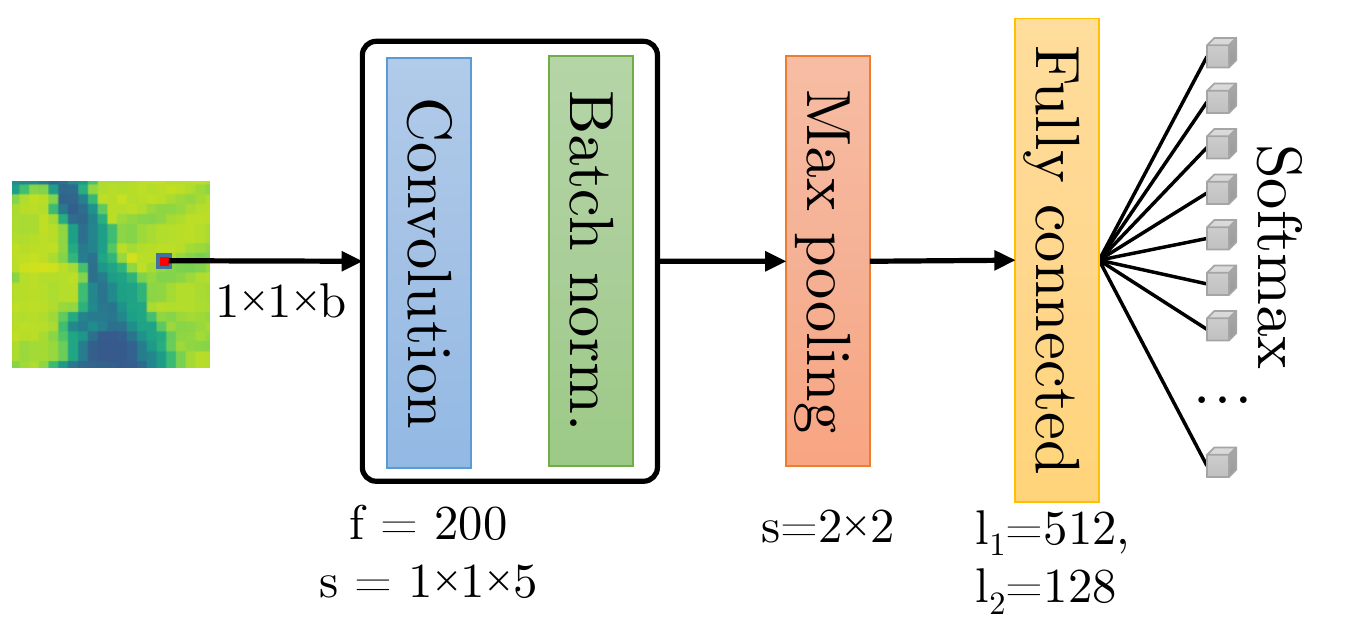}
	\caption{Our 1D network exploits only spectral information with 200 convolutional kernels in the first layer (applied with stride 1). Fully-connected layers (with 512 and 128 neurons) perform multi-class classification.}
	\label{fig:1d_network}
\end{figure}

The main objective of our experimental study was to verify whether a random division of pixels in an input HSI (into $\TrainingSet$ and $\TestSet$) can lead to over-optimistic conclusions about the overall classification performance of the underlying models (we focus on the supervised-learning scenario). In this letter, we exploit two deep-network architectures: a) a recent state-of-the-art 3D CNN (exploiting both spatial and spectral information about the pixels~\cite{Gao_2018}), and our b) 1D convolutional network which benefits from the spectral information only (as in Fig.~\ref{fig:1d_network}, kernels in the convolutional layer are applied in the spectral dimension with stride 1).

In all experiments, we keep the numbers of pixels in $\TrainingSet$ and $\TestSet$ close to those reported in~\cite{Gao_2018} (to ensure fair comparison), and train and validate the deep models using: 1)~balanced $\TrainingSet$ sets containing randomly selected pixels (B), 2)~imbalanced $\TrainingSet$ sets containing randomly selected pixels (IB), and 3)~our patch-based datasets (P), discussed earlier in Section~\ref{sec:method}. For each fold in (3), we repeat the experiments $5\times$, and for (1)~and (2), we perform Monte-Carlo cross-validation with exactly the same number of repetitions (e.g., for Salinas we have 5 folds, each executed $5\times$, therefore we perform $25$ independent Monte-Carlo runs for B and IB). We report per-class, average (AA), and overall accuracy (OA), averaged across all runs.

\begin{table*}[ht!]
\renewcommand{\tabcolsep}{0.13cm}
\centering
\scriptsize
	\caption{Per-class, overall (OA), and average (AA) segmentation accuracy (in \%) for all experimental scenarios (averaged across all executions) obtained for the Salinas Valley dataset (the darker the cell is, the higher accuracy was obtained).}
	\label{tab:results_salinas}
\vrule\pgfplotstabletypeset[%
     color cells={min=0,max=100,textcolor=black},
     /pgfplots/colormap={blackwhite}{ rgb255=(255,255,255) rgb255=(255,170,0)},
    /pgf/number format/fixed,
    /pgf/number format/precision=3,
    col sep=comma,
    columns/Algorithm/.style={reset styles,string type},
    columns/Fold/.style={reset styles,string type},
    columns/Bands/.style={reset styles,string type}%
]{
Algorithm,Fold,C1,C2,C3,C4,C5,C6,C7,C8,C9,C10,C11,C12,C13,C14,C15,C16,OA,AA
3D(P),1,85.69,97.19,0.00,0.00,95.23,0.00,0.00,61.99,94.02,92.09,91.69,0.00,46.72,83.42,78.64,0.06,56.69,51.67
3D(P),2,99.64,99.07,47.88,98.78,0.00,96.16,91.54,96.65,96.66,93.34,92.76,97.70,99.43,96.80,2.92,0.00,72.32,75.58
3D(P),3,99.89,41.27,30.71,99.23,76.12,99.88,97.25,84.03,98.67,67.98,79.37,92.17,99.63,43.02,71.15,77.73,79.80,78.63
3D(P),4,98.12,96.46,43.78,12.73,0.00,100.00,97.90,59.18,6.33,96.35,0.00,93.71,99.07,87.53,90.76,12.79,61.96,62.17
3D(P),5,99.11,41.75,77.06,97.31,87.54,100.00,97.37,72.36,95.05,78.70,93.99,98.85,59.45,0.00,65.53,74.42,77.81,77.40
3D(P),Avg,96.49,75.15,39.89,61.61,51.78,79.21,76.81,74.84,78.14,85.69,71.56,76.49,80.86,62.15,61.80,33.00,69.72,69.09
3D(B),---,99.84,99.30,94.68,99.91,97.51,99.98,99.71,82.82,99.17,96.71,99.42,99.79,99.75,99.73,82.83,99.46,93.04,96.91
3D(IB),---,99.17,99.21,91.52,99.58,97.37,99.97,99.73,90.03,99.66,96.28,96.46,99.52,99.55,99.00,80.63,96.52,94.27,96.51
3D~\cite{Gao_2018},---,100,99.92,99.65,99.78, 99.07,99.97,99.75,94.28,99.97,99.63,99.91,100,100,99.91,97.40,100,98.34,99.33
1D(P),1,94.62,99.23,0.00,0.00,98.50,0.00,0.00,73.93,91.10,91.11,87.11,45.62,98.31,88.52,67.32,0.89,60.65,58.52
1D(P),2,97.84,79.77,58.94,99.32,29.96,99.71,99.41,95.48,94.81,91.54,88.92,99.09,83.73,97.35,4.68,0.00,73.27,76.28
1D(P),3,99.63,70.59,33.63,98.54,98.03,99.91,99.36,82.98,96.75,69.15,89.31,97.68,99.29,83.78,59.53,88.04,83.00,85.39
1D(P),4,97.49,99.83,55.50,98.46,0.00,99.75,99.56,19.37,67.97,92.72,0.00,86.24,97.67,90.55,90.22,18.15,63.90,69.59
1D(P),5,39.99,19.97,20.53,33.29,5.59,98.80,69.61,89.04,8.70,21.01,97.19,26.65,0.00,0.00,3.41,5.65,40.19,33.71
1D(P),Avg,85.91,73.88,33.72,65.92,46.42,79.63,73.59,72.16,71.87,73.11,72.51,71.06,75.80,72.04,45.03,22.54,64.20,64.70
1D(B),---,93.74,94.42,85.00,98.47,81.19,99.53,98.98,59.05,92.49,85.37,90.87,88.94,91.90,92.71,63.33,93.93,80.87,88.12
1D(IB),---,93.78,96.13,76.46,97.68,81.95,99.34,99.21,83.13,97.55,81.00,68.01,93.89,97.72,88.28,43.14,89.93,83.57,86.70

}\vrule
\end{table*}

Our deep models were implemented in \texttt{Python 3.6} with \texttt{Keras}, and we used the ADAM optimizer~\cite{DBLP:journals/corr/KingmaB14} with default parametrization: learning rate of $0.001$, $\beta_1 = 0.9$, and $\beta_2 = 0.999$. The deep-network training terminates if after 15 epochs the accuracy over $\ValidationSet$ (a random subset of $\TrainingSet$) plateaus. In our implementation of 3D CNN, we do not benefit from the \emph{attribute profiles} (APs) proposed in~\cite{Gao_2018}, as the description was not detailed enough to fully reproduce this step. Despite of that, the obtained scores are close to those reported in~\cite{Gao_2018} for Salinas Valley and Pavia University. It is worth noting that the APs are extracted from a whole image before it is split into $\TrainingSet$ and $\TestSet$, which also implies some information leak.

The obtained classification scores are presented in Tables~\ref{tab:results_salinas}--\ref{tab:results_indian_pines} for the Salinas Valley, Pavia University, and Indian Pines datasets, respectively. In addition to the results obtained with our implementation, we also report the scores quoted from~\cite{Gao_2018} (their approach was identical to our B for Pavia University and Salinas Valley, whereas the sets for Indian Pines were imbalanced in~\cite{Gao_2018}). In the Monte-Carlo setting, AA is decreased by up to 6\% for imbalanced sets when compared with their balanced counterparts (for Pavia and Indian Pines) for 1D deep network, and up to 3\% for 3D network (Indian Pines). Also, our implementation renders worse results than those reported in~\cite{Gao_2018} (AA drops by ca. 2.5\%, 6.5\%, and 13\% for Salinas Valley, Pavia University, and Indian Pines, respectively). These differences may be attributed to the lack of APs, which benefit from the global information within an image. Importantly, the differences are much more dramatic when we compare Monte-Carlo cross-validation (both B and IB variants) with our patch-based benchmarks: OA drops by at least 20\% (up to 37\%) for the 3D network, and 6-19\% for the 1D network.

The differences between (B$ -$P) and (IB$-$P) variants are reported in Table~\ref{tab:differences}---they are substantially larger for 3D CNN than for 1D CNN in all cases. Contrary to 1D CNN, in 3D CNN some pixels in $\TestSet$ are seen during training for B and IB variants (as discussed earlier in Section~\ref{sec:intro}), which in our opinion is the only reason why the (B$ -$P) and (IB$-$P) drops in AA and OA are much larger for 3D CNN when compared with 1D CNN. The drop for 1D CNN (as well as for 3D CNN) can also be explained by the fact that our patch-based split results in much more imbalanced representation than IB---in P, we do not ensure including pixels which represent all classes in the training folds generated using the proposed technique, hence we can observe that the deep models are not able to correctly distinguish such classes (e.g., C5, C6, and C7 in the first Pavia University fold). It is a common problem in remote sensing where training examples belonging to all classes of interest may not be available (or are very costly to obtain), hence such scenario is realistic. In addition to that, the patch-based approach extracts pixels located near each other, which may lead to higher correlation between the samples in $\TrainingSet$, therefore affecting the representativeness of the training set $\TrainingSet$.


\begin{table*}[ht!]
\renewcommand{\tabcolsep}{0.13cm}
\centering
\scriptsize
	\caption{Per-class, overall (OA), and average (AA) segmentation accuracy (in \%) for all experimental scenarios (averaged across all executions) obtained for the Pavia University dataset (the darker the cell is, the higher accuracy was obtained).}
	\label{tab:results_pavia}

\vrule\pgfplotstabletypeset[%
     color cells={min=0,max=100,textcolor=black},
     /pgfplots/colormap={blackwhite}{ rgb255=(255,255,255) rgb255=(255,170,0)},
    /pgf/number format/fixed,
    /pgf/number format/precision=3,
    col sep=comma,
    columns/Algorithm/.style={reset styles,string type},
    columns/Fold/.style={reset styles,string type},
    columns/Bands/.style={reset styles,string type}%
]{
Algorithm,Fold,C1,C2,C3,C4,C5,C6,C7,C8,C9,OA,AA
3D(P),1,82.67,98.11,50.12,93.73,0.00,0.00,0.00,79.00,98.77,72.61,55.82
3D(P),2,91.39,68.79,0.00,96.17,99.83,52.67,0.00,93.18,98.77,69.96,66.75
3D(P),3,97.54,95.18,43.86,87.32,99.18,0.00,0.00,55.39,97.97,73.95,64.05
3D(P),4,88.04,93.78,77.22,91.35,0.00,0.00,0.00,41.85,0.00,67.05,43.58
3D(P),5,93.64,53.37,38.41,96.53,99.93,73.33,0.00,81.50,99.65,66.75,70.71
3D(P),Avg,90.66,81.85,41.92,93.02,59.79,25.20,0.00,70.18,79.03,70.07,60.18
3D(B),---,90.22,89.74,90.72,98.53,99.99,87.04,93.55,88.36,99.91,90.59,93.12
3D(IB),---,94.12,98.17,77.00,97.48,99.96,83.69,82.88,87.84,99.58,93.47,91.19
3D~\cite{Gao_2018},---,99.25,99.74,99.76,99.64,100,99.96,98.80,99.48,99.89,99.64,99.61
1D(P),1,94.67,99.20,77.06,84.47,0.00,0.00,0.00,82.28,97.81,75.82,59.50
1D(P),2,94.42,67.69,5.23,95.04,99.52,58.89,0.00,94.57,99.90,70.98,68.36
1D(P),3,96.66,91.24,38.16,89.52,99.65,0.00,0.00,81.28,98.81,74.34,66.15
1D(P),4,90.03,96.22,49.32,69.55,0.00,0.00,0.00,56.48,0.00,66.71,40.18
1D(P),5,91.22,76.65,68.15,95.89,99.88,76.83,0.00,77.66,99.81,78.44,76.23
1D(P),Avg,93.40,86.20,47.58,86.89,59.81,27.14,0.00,78.46,79.27,73.26,62.08
1D(B),---,90.73,86.96,83.09,96.51,99.58,89.59,88.51,80.64,99.86,88.42,90.61
1D(IB),---,93.23,97.43,68.83,87.17,99.40,71.84,64.15,81.35,99.65,89.32,84.78

}\vrule
\end{table*}

\begin{table*}[ht!]
\renewcommand{\tabcolsep}{0.13cm}
\centering
\scriptsize
	\caption{Per-class, overall (OA), and average (AA) segmentation accuracy (in \%) for all experimental scenarios (averaged across all executions) obtained for the Indian Pines dataset (the darker the cell is, the higher accuracy was obtained).}
	\label{tab:results_indian_pines}
\vrule\pgfplotstabletypeset[%
     color cells={min=0,max=100,textcolor=black},
     /pgfplots/colormap={blackwhite}{ rgb255=(255,255,255) rgb255=(255,170,0)},
    /pgf/number format/fixed,
    /pgf/number format/precision=3,
    col sep=comma,
    columns/Algorithm/.style={reset styles,string type},
    columns/Fold/.style={reset styles,string type},
    columns/Bands/.style={reset styles,string type}%
]{
Algorithm,Fold,C1,C2,C3,C4,C5,C6,C7,C8,C9,C10,C11,C12,C13,C14,C15,C16,OA,AA
3D(P),1,0.00,20.83,22.76,9.50,86.28,54.22,0.00,39.53,3.33,56.37,56.44,39.89,70.53,74.88,30.68,0.00,46.83,35.33
3D(P),2,0.00,49.67,41.11,26.15,50.00,64.90,0.00,84.20,3.33,40.16,46.65,0.52,95.70,80.92,40.00,0.00,49.01,38.96
3D(P),3,5.00,37.64,26.00,7.80,24.11,63.32,0.00,83.33,0.00,62.04,45.83,30.70,80.00,80.95,31.79,71.76,50.45,40.64
3D(P),4,15.00,26.67,23.31,28.06,44.88,58.30,0.00,56.92,0.00,53.68,68.17,21.71,17.25,71.30,49.32,80.00,49.28,38.41
3D(P),Avg,5.00,33.70,28.30,17.88,51.32,60.18,0.00,65.99,1.67,53.06,54.27,23.20,65.87,77.01,37.95,37.94,48.89,38.33
3D(B),---,84.06,67.34,84.62,95.98,93.07,98.66,98.75,98.90,100.00,88.91,70.59,77.13,99.55,91.36,61.12,99.77,79.68,88.11
3D(IB),---,62.81,74.81,80.18,83.89,87.67,98.33,76.25,98.53,91.71,84.10,88.85,70.50,99.15,93.92,75.98,99.44,85.90,85.38
3D~\cite{Gao_2018},---,97.83,94.82,97.23, 99.58,99.59,99.59,100,100,100,93.93,97.23,98.99,100,99.76,97.93,98.92,97.57,98.46
1D(P),1,0.00,40.50,41.32,16.20,90.21,96.00,0.00,96.43,24.44,67.44,72.36,62.72,97.74,92.60,41.30,0.00,65.57,52.46
1D(P),2,0.00,73.07,47.78,22.12,75.26,95.87,0.00,90.77,30.00,45.63,75.83,1.31,90.87,95.40,40.79,0.00,64.94,49.04
1D(P),3,31.43,54.06,66.43,31.95,55.89,81.56,0.00,67.29,23.75,70.46,64.71,72.20,93.49,80.39,46.14,94.12,66.42,58.37
1D(P),4,39.29,59.94,50.67,74.82,54.73,95.97,0.00,93.31,0.00,56.69,83.31,38.61,94.49,96.33,45.34,86.05,71.51,60.60
1D(P),Avg,17.68,56.89,51.55,36.27,69.02,92.35,0.00,86.95,19.55,60.05,74.05,43.71,94.15,91.18,43.39,45.04,67.11,55.12
1D(B),---,57.89,67.21,70.01,79.55,85.77,95.35,82.89,97.99,70.53,81.13,60.92,79.78,98.28,95.61,34.57,94.61,73.70,78.26
1D(IB),---,38.35,69.36,63.48,56.19,81.64,92.04,66.17,97.68,43.00,68.29,82.29,62.92,94.95,97.11,52.93,91.32,77.98,72.36

}\vrule
\end{table*}


To shed more light on the statistical importance of the obtained results, we executed two-tailed Wilcoxon tests over AA (for all pairs of the investigated variants). The null hypothesis saying that ``applying different validation strategies and deep models renders the same-quality segmentation'' can be rejected for all cases ($p<0.01$).  It proves the validity of our observations on the over-optimistic performance reported for the 3D deep network (in which spatial information affects pixel's classification) tested using Monte-Carlo cross-validation when compared with our patch-based settings. Although the differences are statistically important for the 1D network as well (B and IB compared with P), they are much smaller than for 3D, which helps understand the importance of avoiding the discussed information leakages.

\begin{table}[ht!]
	\scriptsize
	\centering
	\caption{Average differences (in\%) in OA and AA between different validation settings.}
	\label{tab:differences}
	\renewcommand{\tabcolsep}{0.09cm}
	\begin{tabular}{r|c|cc|ccc}
		\Xhline{2\arrayrulewidth}
        \multicolumn{2}{r|}{Measure$\rightarrow$} & \multicolumn{2}{c|}{OA} & \multicolumn{2}{c}{AA}\\
        \hline
		Dataset$\downarrow$ & CNN & $(\rm B-P)$ & $(\rm IB-P)$ & $(\rm B-P)$ & $(\rm IB-P)$\\
		\hline
        \multirow{2}{*}{Salinas Valley}&1D&16.66&19.36&23.42&22.00\\
        & 3D & 23.32 & 24.55&27.82&27.42\\
        \hline
        \multirow{2}{*}{Pavia University}&1D & 15.16 & 16.07 & 28.52 & 22.70\\
        & 3D & 20.52 & 23.41 & 32.94 & 31.01\\
        \hline
        \multirow{2}{*}{Indian Pines}&1D & 6.59 & 10.87 & 23.14 & 17.24\\
        & 3D & 30.79 & 37.00 & 49.78 & 47.05\\
        \hline
        \multirow{2}{*}{Overall}&1D & 13.25 & 15.76 & 25.16 & 20.89\\
        & 3D & 24.46 & 27.70 & 35.92 & 34.31\\

		\Xhline{2\arrayrulewidth}
	\end{tabular}
\end{table}

\section{Conclusions}\label{sec:conclusions}

In this letter, we showed that the Monte-Carlo cross-validation---which is currently followed in the literature as the strategy to verify emerging hyperspectral image segmentation methods---can very easily lead to over-optimistic conclusions about the performance of segmentation algorithms, because of possible data leaks between training and test sets. We introduced a simple method for creating training-test splits that are free from such shortcomings, and elaborated three ready-to-use hyperspectral image segmentation benchmarks. The experiments, backed up with statistical tests, gave an evidence that our patch-based benchmarks are fairly challenging. The results obtained using two deep-network models (especially the one exploiting the information on pixels' neighborhoods) are much different in the statistical sense (and much worse) compared with Monte-Carlo validation, where the leak of information between the training and test sets occurs. It is especially harmful for algorithms which benefit from the spatial relations concerning a classified pixel, since their classification performance may be seriously over-estimated.

\ifCLASSOPTIONcaptionsoff
  \newpage
\fi

\bibliographystyle{ieeetran}
\bibliography{ref_all}

\begin{thebibliography}{10}
\providecommand{\url}[1]{#1}
\csname url@samestyle\endcsname
\providecommand{\newblock}{\relax}
\providecommand{\bibinfo}[2]{#2}
\providecommand{\BIBentrySTDinterwordspacing}{\spaceskip=0pt\relax}
\providecommand{\BIBentryALTinterwordstretchfactor}{4}
\providecommand{\BIBentryALTinterwordspacing}{\spaceskip=\fontdimen2\font plus
\BIBentryALTinterwordstretchfactor\fontdimen3\font minus
  \fontdimen4\font\relax}
\providecommand{\BIBforeignlanguage}[2]{{%
\expandafter\ifx\csname l@#1\endcsname\relax
\typeout{** WARNING: IEEEtran.bst: No hyphenation pattern has been}%
\typeout{** loaded for the language `#1'. Using the pattern for}%
\typeout{** the default language instead.}%
\else
\language=\csname l@#1\endcsname
\fi
#2}}
\providecommand{\BIBdecl}{\relax}
\BIBdecl

\bibitem{Dundar2018}
T.~Dundar and T.~Ince, ``Sparse representation-based hyperspectral image
  classification using multiscale superpixels and guided filter,'' \emph{IEEE
  GRSL}, pp. 1--5, 2018.

\bibitem{TARABALKA20102367}
Y.~Tarabalka, J.~Chanussot, and J.~Benediktsson, ``Segmentation and
  classification of hyperspectral images using watershed transformation,''
  \emph{Pattern Recognition}, vol.~43, no.~7, pp. 2367 -- 2379, 2010.

\bibitem{Bilgin2011}
G.~Bilgin, S.~Erturk, and T.~Yildirim, ``Segmentation of hyperspectral images
  via subtractive clustering and cluster validation using one-class {SVMs},''
  \emph{IEEE TGRS}, vol.~49, no.~8, pp. 2936--2944, 2011.

\bibitem{Amini2018}
S.~Amini, S.~Homayouni, A.~Safari, and A.~A. Darvishsefat, ``Object-based
  classification of hyperspectral data using random forest algorithm,''
  \emph{Geo-spatial Inf. Sc.}, vol.~21, no.~2, pp. 127--138, 2018.

\bibitem{Li2018}
F.~Li, D.~A. Clausi, L.~Xu, and A.~Wong, ``{ST-IRGS}: A region-based
  self-training algorithm applied to hyperspectral image classification and
  segmentation,'' \emph{IEEE TGRS}, vol.~56, no.~1, pp. 3--16, 2018.

\bibitem{Li2014}
T.~Li, J.~Zhang, and Y.~Zhang, ``Classification of hyperspectral image based on
  deep belief nets,'' in \emph{Proc. IEEE ICIP}, 2014, pp. 5132--5136.

\bibitem{Chen2015}
Y.~Chen, X.~Zhao, and X.~Jia, ``Spectral–spatial classification of
  hyperspectral data based on deep belief network,'' \emph{IEEE J-STARS},
  vol.~8, no.~6, pp. 2381--2392, 2015.

\bibitem{Chen2014}
Y.~Chen, Z.~Lin, X.~Zhao, G.~Wang, and Y.~Gu, ``Deep learning-based
  classification of hyperspectral data,'' \emph{IEEE J-STARS}, vol.~7, no.~6,
  pp. 2094--2107, 2014.

\bibitem{Zhao2016}
W.~Zhao and S.~Du, ``Spectral-spatial feature extraction for hyperspectral
  image classification,'' \emph{IEEE TGRS}, vol.~54, no.~8, pp. 4544--4554,
  2016.

\bibitem{Chen2016}
Y.~Chen, H.~Jiang, C.~Li, X.~Jia, and P.~Ghamisi, ``Deep feature extraction and
  classification of hyperspectral images based on convolutional neural
  networks,'' \emph{IEEE TGRS}, vol.~54, no.~10, pp. 6232--6251, 2016.

\bibitem{Liu2017}
P.~Liu, H.~Zhang, and K.~B. Eom, ``Active deep learning for classification of
  hyperspectral images,'' \emph{IEEE J-STARS}, vol.~10, no.~2, pp. 712--724,
  2017.

\bibitem{Zhong2017}
P.~Zhong, Z.~Gong, S.~Li, and C.~Schönlieb, ``Learning to diversify deep
  belief networks for hyperspectral image classification,'' \emph{IEEE TGRS},
  vol.~55, no.~6, pp. 3516--3530, 2017.

\bibitem{Mou2017}
L.~Mou, P.~Ghamisi, and X.~X. Zhu, ``Deep recurrent neural networks for
  hyperspectral image classification,'' \emph{IEEE TGRS}, vol.~55, no.~7, pp.
  3639--3655, 2017.

\bibitem{Santara2017}
A.~Santara, K.~Mani, P.~Hatwar, A.~Singh, A.~Garg, K.~Padia, and P.~Mitra,
  ``{BASS Net}: Band-adaptive spectral-spatial feature learning neural network
  for hyperspectral image classification,'' \emph{IEEE TGRS}, vol.~55, no.~9,
  pp. 5293--5301, 2017.

\bibitem{Lee2017}
H.~Lee and H.~Kwon, ``Going deeper with contextual {CNN} for hyperspectral
  image classification,'' \emph{IEEE TIP}, vol.~26, no.~10, pp. 4843--4855,
  2017.

\bibitem{Gao_2018}
Q.~Gao, S.~Lim, and X.~Jia, ``Hyperspectral image classification using
  convolutional neural networks and multiple feature learning,'' \emph{Remote
  Sensing}, vol.~10, no.~2, p. 299, 2018.

\bibitem{Ribalta2018}
P.~Ribalta, M.~Marcinkiewicz, and J.~Nalepa, ``Segmentation of hyperspectral
  images using quantized convolutional neural networks,'' in \emph{21st
  Euromicro Conference on Digital System Design}, 2018, pp. 260--267.

\bibitem{RonnebergerFB15}
O.~Ronneberger, P.~Fischer, and T.~Brox, ``U-net: Convolutional networks for
  biomedical image segmentation,'' in \emph{MICCAI}.\hskip 1em plus 0.5em minus
  0.4em\relax Springer, 2015, pp. 234--241.

\bibitem{DBLP:journals/corr/KingmaB14}
D.~P. Kingma and J.~Ba, ``Adam: {A} method for stochastic optimization,''
  \emph{CoRR}, vol. abs/1412.6980, 2014.

\end{thebibliography}

\end{document}